\documentclass[conference,compsoc]{IEEEtran}
\usepackage{amsmath}
\usepackage{hyperref}
\usepackage{amssymb}
\usepackage{graphicx}
\usepackage{enumitem}
\usepackage{booktabs}
\usepackage{multicol}
\usepackage[most]{tcolorbox}
\usepackage{colortbl}
\definecolor{lightgray}{gray}{0.9}

\ifCLASSOPTIONcompsoc
  \usepackage[nocompress]{cite}
\else
  \usepackage{cite}
\fi
\ifCLASSINFOpdf
\else
\fi
\begin{document}
%
\title{SoK: Machine Unlearning for Large Language Models}

\author{
\IEEEauthorblockN{Jie Ren$^{1}$, Yue Xing$^{1}$, Yingqian Cui$^{1}$, Charu C. Aggarwal$^{2}$, Hui Liu$^{1}$}
\IEEEauthorblockA{$^{1}$Michigan State University~~~~~$^{2}$IBM T. J. Watson Research Center\\
\{renjie3,xingyue1,cuiyingq,liuhui7\}@msu.edu~~~~~charu@us.ibm.com
}
}


%


\maketitle

\begin{abstract}
Large language model (LLM) unlearning has become a critical topic in machine learning, aiming to eliminate the influence of specific training data or knowledge without retraining the model from scratch. A variety of techniques have been proposed, including Gradient Ascent, model editing, and re-steering hidden representations. While existing surveys often organize these methods by their technical characteristics, such classifications tend to overlook a more fundamental dimension: the underlying intention of unlearning—whether it seeks to truly remove internal knowledge or merely suppress its behavioral effects.
In this SoK paper, we propose a new taxonomy based on this intention-oriented perspective. Building on this taxonomy, we make three key contributions. First, we revisit recent findings suggesting that many removal methods may functionally behave like suppression, and explore whether true removal is necessary or achievable. Second, we survey existing evaluation strategies, identify limitations in current metrics and benchmarks, and suggest directions for developing more reliable and intention-aligned evaluations. Third, we highlight practical challenges—such as scalability and support for sequential unlearning—that currently hinder the broader deployment of unlearning methods.
In summary, this work offers a comprehensive framework for understanding and advancing unlearning in generative AI, aiming to support future research and guide policy decisions around data removal and privacy.

\end{abstract}


%
\IEEEpeerreviewmaketitle

\section{Introduction}\label{sec:intro}

Large Language Models (LLMs) have demonstrated remarkable capabilities across a wide range of tasks, such as question answering~\cite{kwiatkowski2019natural}, machine translation~\cite{stahlberg2020neural}, text summarization~\cite{el2021automatic}, and dialogue generation~\cite{li2016deep}. A key factor behind this success is the use of web-scale training datasets. However, concerns have been increasingly raised about the inclusion of copyrighted, private, or sensitive information in the training data~\cite{hacker2023regulating, lucchi2024chatgpt}. For example, the New York Times sued OpenAI and Microsoft, claiming that their articles were used in training proprietary LLMs~\cite{nyt2023lawsuit}. Furthermore, legal frameworks such as the General Data Protection Regulation (GDPR) grant individuals the “right to be forgotten”~\cite{voigt2017eu, rosen2011right}, requiring model builders to delete the data upon request.

\begin{figure}[t]
\centering
  \includegraphics[width=\linewidth]{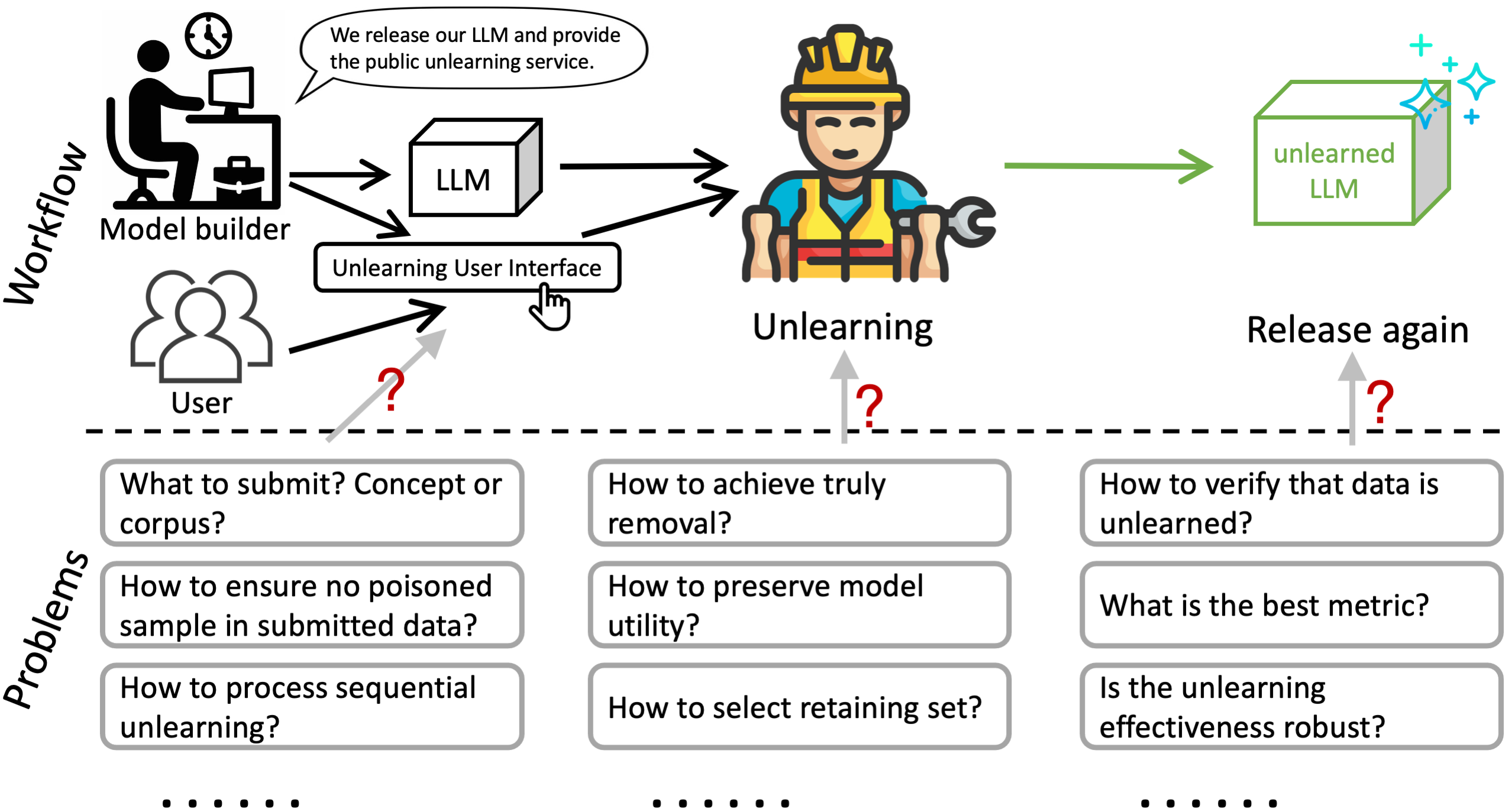}
  \caption{Workflow and existing problems of unlearning}
  \label{fig:intro}
  \vspace{-0.1in}
\end{figure}

To address these concerns, machine unlearning (MU) has been proposed as a technical approach to remove the influence of specific data points from machine learning models without requiring full retraining~\cite{liu2024rethinking, maini2024tofu, ren2024copyright}. Applied to LLMs, the goal of unlearning is to update a model so that its outputs no longer reflect information from a designated set of “forgetting” data, as if the model had never encountered them. 
We illustrate this real-world application scenario in Figure~\ref{fig:intro}: the model developer releases an initial LLM and offers unlearning as a public service. If users suspect that their data were used during training, they can submit unlearning requests. The developer then performs unlearning to address these requests and releases an updated version of the model.

Recent studies have introduced a variety of methods for machine unlearning in large language models (LLMs), such as Gradient Ascent (GA)\cite{yao2023large}, Negative Preference Optimization (NPO)\cite{zhang2024negative}, Representation
Misdirection for Unlearning (RMU)~\cite{liwmdp}. Existing surveys typically classify these methods from a technical perspective—for example, based on whether they rely on fine-tuning~\cite{yao2023large}, use auxiliary models~\cite{huang2024offset}, or leverage In-Context Learning (ICL)~\cite{pawelczyk2023context}. While this implementation-level view is helpful, it often overlooks a deeper distinction: the underlying goal each method seeks to achieve.
Although all methods share the high-level goal of unlearning, their fine-grained intentions can differ substantially. Some aim to truly remove the model’s internal knowledge of the target data~\cite{yao2023large, zhang2024negative}, while others focus on controlling model behavior without removing internal traces~\cite{liwmdp}. We refer to this more nuanced objective as the \textbf{second-level intention} of unlearning.
Based on this perspective, we propose a new taxonomy that categorizes existing methods into \textit{removal-intended} and \textit{suppression-intended unlearning}. This distinction provides clearer insight into the motivation behind each method and helps guide the selection of appropriate techniques for different applications.

Building on this taxonomy, we aim to establish a foundation for understanding unlearning in LLMs and to identify key challenges for advancing the field. To this end, we make three high-level contributions:

\begin{itemize}[leftmargin=1.5em]
    \item We discuss the current understanding of a key question: although the taxonomy suggests that many methods—primarily Gradient Ascent (GA) and its variants—are designed to remove the target knowledge, \textbf{can they truly achieve this goal?} An increasing body of research has begun to scrutinize this assumption. We first summarize the known limitations of GA-based unlearning and introduce a new line of interpretations regarding how GA actually works. We then discuss the inherent challenges of achieving true removal and reflect on whether such complete removal is necessary.
    \item Since effective evaluation plays a crucial role in advancing unlearning research, we present \textbf{a comprehensive overview of evaluation for unlearning}. We outline commonly used metrics and key benchmarks, highlight limitations in current evaluation practices, and suggest potential directions for improvement.
    \item We discuss \textbf{the important gaps that should be considered for practically application in read-world settings}. As shown in Figure~\ref{fig:intro}, these include challenges related to real-world usage scenario (e.g., sequential unlearning requests), minimizing side effects (e.g., preserving the model’s overall capabilities), and verification (e.g., assessing whether the target data has been effectively unlearned), among others.
\end{itemize}

The rest of this SoK paper is organized as follows: Section~\ref{sec:background} reviews the legal foundations related to unlearning. Section~\ref{sec:definition} introduces the problem formulation and key mathematical notations. Sections~\ref{sec:taxonomy}, \ref{sec:removal}, and \ref{sec:suppression} present our proposed taxonomy. Section~\ref{sec:truly} explores the challenge of achieving true removal. In Section~\ref{sec:eval}, we provide a comprehensive review of current evaluation practices and outline future directions. Finally, Section~\ref{sec:gaps} discusses practical gaps, open challenges, and future directions for unlearning.

\section{Background in laws}
\label{sec:background}


The growing interest in unlearning methods for large language models is strongly driven by evolving legal and regulatory frameworks that emphasize \textbf{data privacy}, \textbf{individual rights}, and \textbf{model accountability}. Notably, several national and international laws impose requirements that directly motivate the ability to remove specific training data or knowledge from large language models:

\textbf{United States.} In October 2023, the United States issued Executive Order 14110~\cite{EO14110} to guide the safe and responsible development of artificial intelligence. A key focus of the order is protecting Americans' privacy and civil liberties, particularly as AI makes it easier to extract and act on sensitive personal data. The order emphasizes that the collection and use of such data must remain lawful and secure. It also highlights the need to use policy and technical tools—such as privacy-enhancing technologies—to mitigate risks, including potential infringements on intellectual property and the unauthorized use of copyrighted content generated or processed by AI systems. Notably, the order emphasizes that these techniques should support “\textbf{manageability}” and “\textbf{disassociability}”—the ability to decouple personal data from its computational influence—which directly aligns with the objectives of machine unlearning and underscores the necessity of developing unlearning mechanisms.

\textbf{European Union.} The European Union has implemented a multiple legal frameworks to regulate generative AI, addressing both transparency and privacy concerns. Under Article 53 of the EU AI Act~\cite{EUAIAct2024}, providers of general-purpose AI models are required to comply with Union law on copyright and related rights. They are also required to publish a detailed summary of training data, following a standardized template from the AI Office. In addition, Article 17 of the GDPR~\cite{GDPR2016} provides individuals with the \textbf{right to erasure}, allowing them to request deletion of their personal data when it is no longer necessary, was unlawfully processed, or consent is withdrawn. If the data were made public, controllers must take reasonable technical steps to inform others processing that data of the erasure request. Together, these provisions highlight the growing need for machine unlearning techniques to ensure that AI systems can comply with data deletion and transparency obligations.


In summary, existing data protection laws and emerging AI regulations imply a growing need for unlearning mechanisms to ensure compliance with privacy rights and to maintain transparency and accountability in AI systems.



\section{Problem statement and definitions}
\label{sec:definition}

\begin{table*}[t]
    \centering
   \caption{Taxonomy}
    \resizebox{1\linewidth}{!}{ 
    \begin{tabular}{lllp{7cm}p{0.8cm}p{2cm}}
        \toprule
        Second-level intention & Category & Sub-category & Description & Section & Reference \\
        \midrule 
        Removal-intended & GA-based & GA & GA fine-tunes the model with a reversed standard training loss to negate the training influence of $\mathcal{D}_f$. & \ref{sec:ga}. & \cite{jang2023knowledge, maini2024tofu, shi2024muse} \\\rowcolor{gray!20} 
       \cellcolor{white}  & \cellcolor{white}  & Retaining set & Different retaining set and retaining regularization term is used to maintain the model utility. (Retaining set is also widely used by other methods because of its natural intuition and good compatibility.) & \ref{sec:retaining_set} & \cite{yao2023large, yao2024machine, veldanda2024llm, maini2024tofu, shi2024muse} \\
        & & NPO & NPO gives a new variant whose divergence speed is theoretically proved exponentially slower than GA. & \ref{sec:npo} & \cite{zhang2024negative, fan2024simplicity}  \\\rowcolor{gray!20}
       \cellcolor{white}   & \cellcolor{white} & Gradient conflicts & This type of methods try to mitigate the gradient conflicts between forgetting loss and retaining loss. & \ref{sec:grad_conflict} & \cite{bu2024unlearning, kim2025grail, zhong2025dualoptim} \\
        & & Second-order method & Second-order information provides curvature information in the loss landscape, which allows for more precise and stable updates. & \ref{sec:second-order} & \cite{jia2024soul, huang2025unified, liusophia} \\ \rowcolor{gray!20}
        \cellcolor{white}  & \cellcolor{white} & Selective forgetting & Instead of blindly applying GA to all samples, it first selects the suitable samples, sub-sequences and tokens. & \ref{sec:selective_forgetting} & \cite{buarbulescu2024each, feng2024fine, wang2025selective} \\

        \cmidrule{2-6}

        & Model editing & Task arithmetic & This method extract a task vector from the assistant model and subtract it from the model to unlearn. & \ref{sec:task_ari} & \cite{ilharco2022editing, kim2024negmerge, litask, kuo2025exact} \\
        
        \midrule
        \rowcolor{gray!20}  \cellcolor{white} Suppression-intended & \cellcolor{white} Full-parameter & Fine-grained prob. & Instead of reversing the training loss like GA, this type of methods develops a fine-grained loss to reduce the probability of correct labels. & \ref{sec:fine_prob} & \cite{chatowards, russinovich2025obliviate} \\
        & & Rejection fine-tuning & This method trained the model reject to answering forgetting data using responses like ``I don't know''. & \ref{sec:reject} & \cite{maini2024tofu, liu2024towards, ishibashi2023knowledge, xu2025suv} \\ \rowcolor{gray!20}
        \cellcolor{white}  & \cellcolor{white} & Incorrect labels & This method constructs incorrect labels for forgetting data and fine-tunes with the incorrect labels. & \ref{sec:incorrect_labels} & \cite{eldan2023s, liu2024revisiting, xu2025relearn} \\

        \cmidrule{2-6}
        
        & Input space & ICL-based & Using in-context examples to guide the model for unlearning. & \ref{sec:icul} & \cite{pawelczyk2023context, takashiro2024answer} \\\rowcolor{gray!20}
       \cellcolor{white}   & \cellcolor{white} & RAG-based & A RAG system is used to store the forgetting data. The prompt will retrieve from it to construct a confidentiality instruction. & \ref{sec:rag_based} & \cite{wang2024machine, vilella2025indexing} \\
        & & Agent-based & Agent-based pipelines are used to remove forgetting data at inference time. & \ref{sec:agent_based} & \cite{sanyal2025alu} \\ \rowcolor{gray!20}
        \cellcolor{white}  & \cellcolor{white} & Embedding corruption & The input embeddings of forgetting data tokens are changed to suppression forgetting data. & \ref{sec:embedding_corruption} & \cite{liu2024large} \\

        \cmidrule{2-6}

        & Representation & Re-steering & The hidden representations related to forgetting data is re-steered to random vectors or a rejeciton area. & \ref{sec:resteering} & \cite{liwmdp, huu2024effects, shen2025lunar, huu2025improving, hu2025falcon, wang2024large} \\ \rowcolor{gray!20}
       \cellcolor{white}  & \cellcolor{white} & Editing by SAE & SAE interprets the features in hidden representations. Thie type of methods unlearn by negating the SAE features related to forgetting data. & \ref{sec:sae} & \cite{farrell2024applying, khoriaty2025don, li2025sauce, muhamed2025saes} \\
        & & Additional modules & This kind of method freezes the original model parameters, and use a play-and-plug module to change the representation. & \ref{sec:additional_module} & \cite{chen2023unlearn, ren2025general, gao2025large, hu2022lora} \\\rowcolor{gray!20}
        \cellcolor{white} & \cellcolor{white} & Localization & These methods locate the parameters related to forgetting data and unlearn by techniques such like pruning. & \ref{sec:localization} & \cite{pochinkov2024dissecting, liu2025modality} \\

        \cmidrule{2-6}

        & Output space & Logits difference & Logits-difference methods subtract or offset output logits using assistant models trained to isolate the influence of forgetting data. & \ref{sec:logits_diff} & \cite{huang2024offset, ji2024reversing} \\ \rowcolor{gray!20}
       \cellcolor{white}  & \cellcolor{white} & Retrieval-based & This method retrieves forgetting data suppresses forgetting content by blocking semantically matched answer tokens. & \ref{sec:retrieval_based} & \cite{deng2025guard} \\
        
        \bottomrule
    \end{tabular}
    }
    \label{tab:taxo}
\end{table*}

In this section, we provide the mathematical definitions of unlearning problem and necessary concepts in LLM.

We consider a language model $f_\theta$ with $\theta$ as the model parameters.
Let $x$ denote an input (e.g., a prompt or query), and $y = f_\theta(x) = (y_1, y_2, \dots, y_T)$ denote an output sequence generated in response to $x$ using the model $f$.
Let $P_\theta(y \mid x)$ denote the conditional probability assigned by the model to output $y$ given input $x$.
For brevity, we write 
\begin{align}
    p = P_\theta(y \mid x)
    \label{eq:output_dis}
\end{align}
to represent the model’s output distribution under $\theta$.

In the problem of LLM unlearning, we define the output distribution of three different models as follows:

(1) $p_0$ represents the output distribution of the original model trained on the full dataset;

(2) $p^{*}$ represents the output distribution of the ideally unlearned model that is retrained on the dataset without forgetting data;

(3) $p_{u}$ represents the output distribution of the unlearned model, i.e., after attempting to forget certain data from $p_0$.

We give the general definition of LLM unlearning as follows:

Given $f$ and the forgetting set $\mathcal{D}_{f}$, the task of unlearning is to find a $p_{u}$ that can that can approximate the ideally unlearned distribution $p^{*}$ as much as possible. Formally, if we use $D(p_u, p^*)$ to denotes a general divergence or distance between the two distributions,
the task of unlearning can be formulated as
\begin{align}
    p_u = \mathop{\arg \min}\limits_{p\in\mathcal{P}} D(p, p^*),
    \label{eq:define_unlearn}
\end{align}
where $\mathcal{P}$ represents the set of candidate output distributions determined by the specific unlearning algorithm.

Eq.~\eqref{eq:define_unlearn} has two key implications. First, the unlearned model should provide information of the forgetting data as minimal as possible. Second, it should retain all other knowledge, preserving the ability to respond accurately to any input $x \notin \mathcal{D}_{f}$. This preservation of general performance is referred to as the model’s \textbf{\textit{utility}} in the context of LLM unlearning.

To obtain $p_u$, unlearning methods are not necessarily limited to adjusting model parameters $\theta$. For example, In-Context Unlearning (ICUL)~\cite{pawelczyk2023context} and RAG-based unlearning~\cite{wang2024machine} modify the input in order to suppress the model's learned response:
\begin{align}
    p_u = P_\theta(y \mid x, x_0),
\end{align}
where $x_0$ represents an auxiliary input—such as an in-context example used in ICUL or a modified prompt generated via retrieval-augmented generation (RAG)—designed to guide the model toward forgetting the target information.
\section{Removal or suppression: two paths in unlearning}
\label{sec:taxonomy}

As noted in Section~\ref{sec:intro}, prior surveys have predominantly focused on the technical mechanisms of unlearning methods, often overlooking a more fundamental consideration: the underlying intention each method is designed to fulfill. While many approaches share similar architectures or loss functions, their goals may differ substantially in terms of how they address the presence of the forget set within the model.
Recognizing these second-level intentions is essential. Without an intention-aware taxonomy, it becomes difficult to assess whether an unlearning method aligns with the expectations of regulators, end-users, or downstream applications.

In this section, we propose a new taxonomy grounded in these intentions. We classify existing unlearning methods into two primary categories:

\begin{itemize}
    \item \textbf{Removal-intended unlearning}, which aims to genuinely eliminate the model’s internal knowledge or training trace of the forget set;
    \item \textbf{Suppression-intended unlearning}, which accepts that the model may still encode the forgotten data, but restricts its output to behave as if it had been forgotten.
\end{itemize}

Understanding the intention behind an unlearning method is crucial for evaluating its appropriateness in different real-world scenarios. For example, legal compliance with data protection laws may demand strong guarantees of removal, whereas applications like harmful knowledge filtering may only require output-level suppression. A taxonomy rooted in intention thus helps bridge the gap between technical design and application-specific goals.

Table~\ref{tab:taxo} provides an overview of our proposed taxonomy. For removal-intended unlearning, we divide existing methods into two main branches: GA-based approaches and model editing techniques. GA-based methods represent the core of this category, and we further introduce their variants as subcategories.
For suppression-intended unlearning, rather than following prior taxonomies that primarily focus on technical mechanisms, we instead structure our analysis around the specific components of LLMs that each method targets—such as the input space, hidden representations, or output tokens. For each component, we identify representative methods, discuss their suitability, and highlight the technical challenges and design considerations involved.

In the next two sections, we provide a detailed discussion of each category within this intention-based taxonomy.


\section{Removal-intended unlearning}
\label{sec:removal}

Removing specific knowledge or the influence of particular data is the ideal goal of unlearning~\cite{bourtoule2021machine}. Achieving this goal inevitably requires modifying the model parameters $\theta$, as the learned knowledge and data influence are encoded within them. Gradient Ascent (GA)~\cite{jang2023knowledge} and its variants~\cite{yao2023large,zhang2024negative,liu2022continual} have played an important role in removal-intended unlearning because of its straightforward intuition of reversing the training process. Besides them, a few works also propose to directly edit the model parameters such as Task Arithmetic~\cite{ilharco2022editing}. 
In this subsection, we majorly introduce the works that intend to use GA and its variants, and some minority work relies on other methods.


\subsection{Gradient Ascent and its variants}

\subsubsection{Gradient Ascent}
\label{sec:ga}
The idea of GA stems from the machine unlearning for classification models~\cite{graves2021amnesiac, thudi2022unrolling}, and is inherited by the unlearning for LLM. 
The core intuition is that \textbf{by fine-tuning with a reversed training loss of LLM, GA can negate the training influence of} $\mathcal{D}_f$.
The formulation of GA can be denoted as:
\begin{align}
    \mathcal{L}_{\mathrm{GA}} 
    &= - \mathcal{L}_{\text{train}} = {E}_{(x, y) \sim \mathcal{D}_f} \left[ \log \pi_{\theta}(y \mid x) \right],
    \label{eq:ga_init}
\end{align}
where $\mathcal{L}_{\text{train}}$ denotes the standard next-token prediction loss for LLM training, and $\pi_{\theta}(y \mid x)$ represents the model's likelihood (or predictive probability) of generating target $y$ given input $x$. By fine-tuning to minimize $\mathcal{L}_{\mathrm{GA}}$, the model reduces the likelihood of generating $y$. This simple yet effective idea has shown strong performance in unlearning tasks~\cite{jang2023knowledge, maini2024tofu, shi2024muse}. However, GA is highly sensitive to hyperparameters~\cite{liu2024rethinking} and can lead to significant utility degradation~\cite{zhang2024negative}. These limitations have motivated the development of various GA-based variants aimed at improving model utility. In the following, we introduce several such variants, each driven by different design motivations.

\subsubsection{Retaining set}\label{sec:retaining_set} Researchers think that the reversed loss used in GA-based unlearning may degrade model utility due to its opposite optimization direction~\cite{yao2023large, yao2024machine}. To mitigate this issue, a natural solution is to introduce \textbf{retaining set} $\mathcal{D}_r$ to constrain the distortion caused by the reversed loss. This leads to a variant of GA in which the fine-tuning objective is split into two components~\cite{yao2024machine, yao2023large, veldanda2024llm}:
\begin{align}
    \mathcal{L}_{\mathrm{u}} = \mathcal{L}_{f} + \lambda \mathcal{L}_r = \mathcal{L}_{\mathrm{GA}} + \lambda \mathcal{L}_r,
\end{align}
where $\mathcal{L}{\mathrm{u}}$ denotes the overall unlearning loss, composed of a forgetting loss $\mathcal{L}{f}$ and a retaining loss $\mathcal{L}_r$, with $\lambda$ controlling the strength of the retaining term. There are typically two choices for defining $\mathcal{L}r$. The first is standard gradient descent on the retaining set $\mathcal{D}r$, leading to what is known as the Gradient Difference method~\cite{maini2024tofu, shi2024muse}. The second option uses the Kullback–Leibler (KL) divergence between the output distributions of the updated (unlearned) model and a reference (pre-unlearning) model:
\begin{align*}
    \mathcal{L}_r = \underset{(x, y) \sim \mathcal{D}_r}{E} \sum_{i=1}^{|y|} \left[\mathrm{D}_{KL}\left(\mathrm{p}_\theta(\cdot \mid x, y_{<i}) \| \mathrm{p}_{\mathrm{ref}}(\cdot \mid x, y_{<i})\right)\right],
\end{align*}
where $\mathrm{D}_{KL}\left(\cdot \mid \cdot\right)$ is the KL divergence, and $\mathrm{p}(\cdot \mid x, y_{<i})$ is the next-token probability distribution of the updated model and the reference model~\cite{yao2023large, veldanda2024llm}.
Due to its natural intuition and good compatibility (only adding one loss term), retaining set is also widely used by the variants of GA and fine-tuning-based suppression-intended unlearning.

\subsubsection{Negative Preference Optimization}\label{sec:npo} NPO is a notable variant of GA, distinguished by its strong ability to preserve model utility~\cite{zhang2024negative, fan2024simplicity}. NPO points out that GA can rapidly reduce the model utility to a catastrophic collapse because of the ``\textit{divergent} nature of the gradient ascent algorithm due to the fact that it
maximizes the standard next-token prediction loss''~\cite{zhang2024negative}. Thus, to slow down the divergence, they propose NPO which is denoted as:
\begin{align*}
    \mathcal{L}_{\mathrm{NPO}} = -\frac{2}{\beta} {E}_{(x, y) \sim \mathcal{D}_f} \left[ \log \sigma\left(-\beta \log \frac{\pi_\theta(y \mid x)}{\pi_{\mathrm{ref}}(y \mid x)}\right) \right],
\end{align*}
where $\sigma(t)=1 /\left(1+e^{-t}\right)$ is the sigmoid function, and $\beta > 0$ is the inverse temperature. In~\cite{zhang2024negative}, the divergence speed is theoretically proved exponentially slower than GA. 


\subsubsection{Mitigating gradient conflicts}\label{sec:grad_conflict} Although incorporating the retaining loss $\mathcal{L}_r$ helps mitigate the side effects of the reversed next-token prediction loss, jointly optimizing two objectives with opposing goals can lead to gradient conflicts. Several methods have been proposed to address this challenge from different angles~\cite{bu2024unlearning, kim2025grail, zhong2025dualoptim}.

Bu et al.~\cite{bu2024unlearning} construct a dynamic update direction by normalizing and contrasting the gradients of the retaining and forgetting objectives:
\begin{align*}
    g_{\mathrm{NGDiff}}=\frac{g_r}{\left\|g_r\right\|}-\frac{g_f}{\left\|g_f\right\|},
\end{align*}
where $g_r$ and $g_f$ are the gradients from retaining and forgetting term.
This direction is positively correlated to $g_r$ and negatively correlated to $g_f$ (see Lemma 4 in \cite{bu2024unlearning}). Thus, each update reliably decreases $\mathcal{L}_r$ while increasing $\mathcal{L}_f$.

Kim et al.~\cite{kim2025grail} tackle the problem by partitioning model parameters based on their relevance to forgetting or retaining targets. Using gradient-based analysis, they freeze utility-critical parameters and apply gradient ascent to forget-related ones and gradient descent to other parameters to reinforce retained knowledge.

Zhong et al.~\cite{zhong2025dualoptim} observe that gradient conflicts often stem from momentum. To address this, they use separate optimizers for $\mathcal{L}_f$ and $\mathcal{L}_r$, isolating momentum updates and reducing interference for more stable optimization.

\subsubsection{Second-order information}\label{sec:second-order} Second-order updates can more effectively adjust model parameters by accounting for curvature information in the loss landscape. This allows for more precise and stable updates, especially when balancing objectives between forgetting and retaining. 

Jia et al.\cite{jia2024soul} view unlearning as a second-order adjustment inspired by influence functions. Using the Sophia optimizer\cite{liusophia}, they efficiently approximate the Hessian, enabling scalable and robust unlearning across diverse loss functions and models.

Huang et al.~\cite{huang2025unified} leverage second-order information to capture the curvature of the loss landscape, enabling the model to identify and preserve important parameters that contribute most to performance on retained data. They constrain updates within a remain-preserving manifold, minimizing output distortion and reducing utility loss.

\subsubsection{Selective forgetting} 
\label{sec:selective_forgetting}
We include selective forgetting~\cite{buarbulescu2024each, feng2024fine, wang2025selective} as the final category of GA variants, as it reflects a more critical understanding of GA, which is an important step forward in the development of LLM unlearning. Instead of blindly applying GA to all samples, it first identifies and selects the most suitable samples and tokens for unlearning.

Not all tokens in the forgetting set $\mathcal{D}_f$ are equally relevant to the target knowledge. However, as shown in Eq.\eqref{eq:ga_init}, Gradient Ascent treats every token in $y$ uniformly during unlearning, which is suboptimal. For instance, to forget the knowledge of ``\textit{Watermelon on the Moon?}", a sequence like ``\textit{The author of Watermelon on the Moon was born in 1988}" may exist in $\mathcal{D}_f$. Applying GA to unrelated tokens such as ``\textit{The author of}" or ``\textit{was born in}" unnecessarily harms utility. This highlights a limitation of GA, which recent works aim to address at different levels\cite{buarbulescu2024each, feng2024fine, wang2025selective}.

Barbulescu et al.~\cite{buarbulescu2024each} handle this problem at sample level. They propose a memorization score based on the ratio of memorized tokens. Their method adaptively selects and unlearns only the highly memorized samples in each iteration, thereby reducing privacy risk from memorization outliers. Wang et al.~\cite{wang2025selective} refine this idea at the token-span level, selecting low-probability tokens online as they propose that the low probability tokens in a sequence might contain more sensitive information. Feng et al.~\cite{feng2024fine} take a direct approach by scoring each token with a trained detector and applying GA only to the highest-scoring ones.



\subsection{Model editing by task arithmetic} \label{sec:task_ari}

Task arithmetic is a model editing method applied in LLM unlearning~\cite{ilharco2022editing, kim2024negmerge, litask, kuo2025exact}. It introduces the task vector, defined as the weight difference between a fine-tuned model $\theta^{\prime}T$ and its pre-trained counterpart $\theta^{(0)}$ on task $T$:
\begin{equation}
\Delta \theta_{T} = \theta^{\prime}_{T} - \theta^{(0)}.
\end{equation}
Unlearning is achieved by subtracting this vector:
\begin{equation}
\theta = \theta^{(0)} - \lambda \Delta \theta_{T}, \quad \lambda > 0.
\end{equation}
The method is parameter-efficient and theoretically effective when the unlearned task is either irrelevant or contradictory to the retained tasks.

Kim et al.~\cite{kim2024negmerge} aggregate multiple task vectors from models fine-tuned with diverse hyperparameters. Instead of selecting a single task vector, it merges only the elements with consistent signs across all task vectors—presumed to reflect forget-related knowledge—while masking conflicting elements. This merged vector is then subtracted from the original model to perform unlearning.

\section{Suppression-intended unlearning}
\label{sec:suppression}
Thoroughly removing knowledge from a model remains a challenge task. Technically, true removal often requires modifying all parameters, which is difficult to scale and may harm utility. Conceptually, fully and irreversibly erasing the forgetting data is hard to guarantee (see Section~\ref{sec:truly}). As a result, suppression-intended methods offer a more pragmatic alternative: instead of eliminating traces of the forgotten data, they aim to prevent the model from generating outputs that could expose sensitive or copyrighted content. While removal-intended methods—typically using GA—focus on changing model parameters $\theta$, suppression methods can intervene at any stage of the generation pipeline in Eq.~\ref{eq:output_dis}, including the input $x$, hidden state $h(x)$, or output logits $l(x)$. This flexibility has led to diverse suppression strategies. In this section, we categorize them based on the component they target, highlighting each method’s motivations and its alignment with the properties of that component.


\subsection{Full-parameter fine-tuning} Before diving into the different components, we first introduce several full-parameter fine-tuning methods. Although these methods update the entire set of model parameters—similar to removal-intended unlearning—they are possible to be less detrimental to model utility, as they do not rely on reversed training loss. The methods share a key motivation: although GA can reduce the likelihood of generating forgetting data, it does not tell the model what to output after unlearning, which might reduce of model utility. These approaches aim to mitigate such side effects.

\subsubsection{Fine-grained probability}
\label{sec:fine_prob}
Fine-grained probability is closely tied to GA-based methods~\cite{chatowards, russinovich2025obliviate}, which push the model to generate nonsensical responses to minimize $\mathcal{L}_{\mathrm{GA}}$ in Eq.\ref{eq:ga_init}. This also explains the divergent property of GA. We only want the answer to change a little to hide the forgotten data, but GA often leads the model to generate entirely incoherent outputs. To address this, Cha et al.~\cite{chatowards} introduce a new Inverted Hinge Loss (IHL): 
\begin{align*}
    \mathcal{L}_{\mathrm{IHL}}(x,y)=1+\mathrm{p}_\theta\left(y_t \mid x,y_{<t}\right)-\max _{v \neq y_t} \mathrm{p}_\theta\left(v \mid x,y_{<t}\right)
\end{align*}
IHL reduces the target token's probability while promoting its closest alternative token, making unlearning more controlled, efficient, and less damaging to overall model utility. Russinovich et al.~\cite{russinovich2025obliviate} propose a forget loss that applies KL-divergence between the original token distribution and a modified distribution with the target token removed:
\begin{align*}
    \mathcal{L}_{f} = \mathrm{D}_{K L}\left( 
\mathrm{softmax}(l) \,\big\|\, \mathrm{softmax}(l \backslash x_{\text {target }}) 
\right),
\end{align*}
where $l \in \mathbb{R}^k$ be the logits of the model's top-$k$ predictions at a target position, and let $x_{\text{target}}$ be the top-1 predicted token to be unmemorized. This pushes the model away from memorized outputs while preserving fluency.

\subsubsection{Rejection fine-tuning}
\label{sec:reject}
Rejection fine-tuning trains the model to respond to forgetting data with a fixed rejection template (e.g., “I don’t know”)\cite{maini2024tofu, liu2024towards, ishibashi2023knowledge, xu2025suv}. This avoids reversed losses and is based on preference optimization\cite{rafailov2024direct}.

\subsubsection{Incorrect labels}
\label{sec:incorrect_labels}
This strategy provides plausible but incorrect responses (e.g., wrong names or facts)~\cite{eldan2023s, liu2024revisiting, xu2025relearn}. Unlike explicit rejection, it mimics a model unfamiliar with the forgetting data, producing more natural yet misleading outputs that help mask unlearning traces.

\subsection{Input space} Unlearning in the input space best preserves model utility, as it avoids modifying model parameters or the inference process—only the input $x$ is altered. This subsection covers approaches based on In-Context Learning (ICL)~\cite{pawelczykcontext, takashiro2024answer}, Retrieval-Augmented Generation (RAG)~\cite{wang2024machine, vilella2025indexing}, LLM agent~\cite{sanyal2025alu}, and embedding corruptions~\cite{liu2024large}.

\subsubsection{In-Context Unlearning (ICUL)}
\label{sec:icul}
ICUL is a counterpart of ICL. It operates by combining two types of examples: \textit{Label Flipping}, where the labels of target instances are randomly altered, and \textit{Context Padding}, where correctly labeled examples are added to stabilize the prompt. ICUL has good performance and is efficient and compatible with black-box LLMs~\cite{pawelczyk2023context}. Another ICL-based method~\cite{takashiro2024answer} first fine-tune the model to learn the unlearning task in ICL. 

\subsubsection{RAG-based unlearning}
\label{sec:rag_based}
RAG-based unlearning~\cite{wang2024machine} stores forgetting data in a retrieval system. When a user prompt is received, the system checks if any forgetting content is retrieved. If so, it appends a confidentiality instruction to the retrieved knowledge, prompting the LLM to avoid generating related outputs. Despite adding a retrieval step, this approach preserves model utility well and offers a promising alternative to direct model-level forgetting~\cite{vilella2025indexing}. It also naturally supports sequential unlearning, accommodating ongoing unlearning requests. To reduce latency, we suggest a parallel design: vanilla generation and retrieval run simultaneously. Since most prompts don't involve forgetting data, the retrieval is empty and their responses can be returned directly. If retrieval is non-empty, a second pass with suppression is triggered—ensuring fast response for benign prompts.

\subsubsection{Agent-based unlearning}
\label{sec:agent_based}
ALU (Agentic LLM Unlearning)~\cite{sanyal2025alu} introduces a training-free, model-agnostic approach to LLM unlearning via a four-agent pipeline: the Vanilla Agent generates the initial response, the AuditErase Agent produces multiple redacted versions, the Critic Agent scores these versions for unlearning efficacy and utility, and the Composer Agent synthesizes a final high-quality, sanitized output. This agentic framework supports dynamic unlearning requests, demonstrates strong robustness to adversarial prompts, and scales to thousands of targets without degrading performance. However, ALU incurs more inference-time computation than typical post hoc unlearning methods due to multiple LLM-agent calls.

\subsubsection{Embedding Corruption}
\label{sec:embedding_corruption}
Beyond modifying the input $x$, Embedding-Corrupted Prompts is proposed to change its embeddings~\cite{liu2024large}. It uses a trained prompt classifier to detect whether an input query falls within the forgetting scope, and applies learned perturbations to the prompt embeddings at inference time. These perturbations are optimized offline to align the model's output with that of a retained model that has never seen the forgetting data.

\subsection{Hidden representation space} Hidden representations in LLMs encode rich semantic information, offering a broad space for unlearning techniques to explore. We categorize them into three main types: (1) Re-steering hidden representation~\cite{liwmdp, huu2025improving, shen2025lunar, huu2024effects, hu2025falcon, wang2024large}, (2) Editing by Sparse Autoencoder~\cite{farrell2024applying, khoriaty2025don, li2025sauce, muhamed2025saes}, pruning~\cite{pochinkov2024dissecting, liu2025modality} and (3) Additional modules~\cite{chen2023unlearn, ren2025general, li2025effective, gao2025large}.

\subsubsection{Re-steering hidden representation}
\label{sec:resteering} This kind of method suppresses the semantic information of forgetting data in the representation space, especially when such knowledge appears in hidden states.

Representation Misdirection for Unlearning (RMU)~\cite{liwmdp} is a representative method in this class. It fine-tunes a consecutive Transformer layers to map the forgetting data representation into a random vector. The objective is:
\begin{align}
    \mathcal{L}_{RMU}= & ~ E_{x_f \in \mathcal{D}_{f}}\left\|h_\theta^{(l)}\left(x_f\right)-c u\right\|_2^2 \nonumber \\
    + & ~\alpha E_{x_r \in \mathcal{D}_{r}}\left\|h_\theta^{(l)}\left(x_r\right)-h_{\mathrm{ref}}^{(l)}\left(x_r\right)\right\|_2^2,
    \label{eq:rmu}
\end{align}
where $u$ is a fixed random unit vector sampled from a uniform distribution $\mathcal{U}(0,1)$, which serves as the misdirection target, and $c$ is a coefficient controlling the magnitude of the forget-target representation (i.e., the norm of the vector $cu$). $h_\theta^{(l)}(x)$ and $h_{\mathrm{ref}}^{(l)}(x)$ are the $l$-th layer representations of $x$ of unlearned model and reference model (model before unlearning). The first term suppresses the forgetting data by mapping its representation to a random vector, while the second term preserves performance on retaining data.

Dang et al.~\cite{huu2024effects} observe that RMU fails to converge in deeper layers due to larger representation norms. This is because the representation norms increase in deeper layers. Forcing these high-norm vectors toward a fixed small-norm target $cu$ is unnatural and leads to instability, requiring extensive tuning of $c$. To fix this, they scale the target vector using the original norm of each sample:
\begin{align*}
    E_{x_f \in \mathcal{D}_{f}}\left\|h_\theta^{(l)}\left(x_f\right)-\beta\| h_{{\text {ref }}}^{(l)}\left(x_f\right)\|\cdot u \right\|_2^2.
\end{align*}
This norm-matching approach improves convergence and allows RMU to work effectively across all layers.


Similar to RMU, Shen et al.~\cite{shen2025lunar} propose to re-steer the representations of forgetting data into a known refusal region. This region is defined by the representations of prompts that naturally cause the model to refuse, such as harmful or unethical queries (e.g., jailbreak triggers), and fictitious or nonsensical questions (e.g., ``\textit{What’s the capital of \$7\&a\#!}'').

Hu et al.\cite{hu2025falcon} propose a method that combines hidden representation re-steering with gradient conflict mitigation and second-order optimization. It selects minimally entangled parameters using mutual information, separates forget and retain data via contrastive learning, and pushes forgetting data away from the reference model's representations. Gradient projection resolves optimization conflicts, while the Sophia optimizer\cite{liusophia} ensures stable convergence.


\subsubsection{Editing by SAE} \label{sec:sae} Recently, Sparse Autoencoder has been used as an important tool in interpretability~\cite{cunningham2023sparse, shu2025survey}, and are now being explored for unlearning by disentangling and removing features of the forgetting data from the representation space~\cite{farrell2024applying, khoriaty2025don, li2025sauce, muhamed2025saes}.

We begin by introducing how Sparse Autoencoders (SAEs) support interpretability. Given an input $\mathrm{x}$—typically a hidden-layer representation $h^{(l)}(x)$ in LLMs—SAE learns a sparse code $z$ by minimizing the reconstruction loss with a sparsity penalty~\cite{ng2011sparse}:
\begin{align}
    \mathcal{L}_{\mathrm{SAE}}=\|\mathrm{x}-\hat{\mathrm{x}}\|_2^2+\lambda \sum_{i=1}^d \rho\left(z_i\right),
\end{align}
where the encoder maps $\mathrm{x}$ to $z$, and the decoder reconstructs $\hat{\mathrm{x}}$ from $z$. The regularizer $\rho(z_i)$ (e.g., KL divergence) encourages most neurons to remain inactive, so each neuron in $z$ captures distinct, human-interpretable patterns. Each hidden neuron in $z$ activates only in response to specific, meaningful input features. 

Farrell et al.~\cite{farrell2024applying} propose to train such an SAE for the model to unlearn with the hidden representation $h^{(l)}(x)$ of $l$-th layer as the input of SAE. By measuring feature activations on forgetting data (e.g., harmful bio-weapons in WMDP\cite{liwmdp}), they identify SAE features {$z^f_i$} associated with the forget set—those frequently activated in the forget set but not in the retain set. During inference, unlearning is achieved by suppressing these features, setting $z^f_i = -\mathrm{C}$ ($\mathrm{C} > 0$). The modified vector is then decoded into $\hat{h}^{(l)}(x)'$ to replace the original representation, weakening the model’s ability to generate related content.


Khoriaty et al.~\cite{khoriaty2025don} further find a ``refusal feature'' that can safely return a refusal message.
Li et al.~\cite{li2025sauce} extend this method to vision-language model.

\subsubsection{Additional modules} \label{sec:additional_module} This type of method typically freezes the original model parameters and introduces a plug-and-play module to alter the representation. LoRA~\cite{hu2022lora} is a common approach in suppression-intended unlearning. Regardless of whether the loss is removal- or suppression-intended, LoRA freezes the base model, meaning it cannot truly erase learned knowledge. Beyond LoRA, several lightweight modules have been proposed specifically for LLM unlearning. These modules add minimal inference overhead while effectively suppressing the influence of forgetting data. Below, we introduce four such approaches.

Chen et al.~\cite{chen2023unlearn} propose a method that inserts lightweight unlearning layers into transformers. These layers are trained using an objective that preserves performance on retained data while forgetting target data using KD divergence. The forgetting loss is: 
\begin{align*}
    \mathcal{L}_f= - \underset{(x, y) \sim \mathcal{D}_f}{E} \sum_{i=1}^{|y|}\left[\mathrm{D}_{K L}\left(\mathrm{p}_\theta\left(\cdot \mid x, y_{<i}\right) \| \mathrm{p}_{\theta,l}\left(\cdot \mid x, y_{<i}\right)\right)\right],
\end{align*}
where $\mathrm{p}_{\theta,l}$ is the next-token output probability distribution of the model with unlearning layers. This method can efficiently handle sequential deletion by merging multiple unlearning layers into one via linear regression, enabling scalable and composable unlearning without accessing the original training data.

Ren et al.~\cite{ren2025general} propose a Gated Representation Unlearning (GRUN) which uses a gate function to control the unlearning strength. GRUN is based on Representation Fine-tuning (ReFT) by adding a gate on ReFT. ReFT changes the $l$-th layer's output representation in a residual form. The changed representation is
\begin{align}
    h_{\text{new}}^{(l)}(x)=h^{(l)}(x)+\phi\left(h^{(l)}(x)\right),
\end{align}
where $\phi$ is a trainable low-rank linear transformation.
GRUN uses a soft gate function $g$ to control the unlearning strength:
\begin{align}
    h_{\text{new}}^{(l)}(x)=h^{(l)}(x)+g\left(h^{(l)}(x)\right)\phi\left(h^{(l)}(x)\right),
\end{align}
where $g$ is a single-output regression model (linear regression or Multi-Layer Perceptron neural network) with a sigmoid function following the output. The advantage of GRUN is that the soft gate function can be close to zero (i.e. $h_{\text{new}}^{(l)}(x) \approx h^{(l)}(x)$) when retaining data is input, which large preserve the model utility.

Gao et al.~\cite{gao2025large} use a similar idea for LoRA. They use a detector to measure the similarity between input and unlearning data. The similarity score is used to control the loading of LoRA and then the strength of unlearning.

\subsubsection{Localization} \label{sec:localization} A few works propose to localize the parameters related to forgetting data~\cite{guo2024mechanistic, yang2025faithun} and then unlearning by methods such as neural network pruning identifies neurons~\cite{pochinkov2024dissecting, liu2025modality}. However, a recent paper challenges the core assumption of localized unlearning~\cite{lee2025does}—that effective knowledge removal in LLMs requires updates to specific parameter regions. Through controlled experiments, the authors demonstrate that unlearning can be achieved equally well by modifying randomly selected regions, suggesting that no fixed set of parameters is necessary for forgetting. These findings call into question the causal link between parameter localization and unlearning success.

\subsection{Output space} In this section, we introduc the methods focusing on output space which contains the modification on the logits and retrieval-based methods.

\subsubsection{Logits difference} \label{sec:logits_diff} Huang et al.~\cite{huang2024offset} propose $\delta$-unlearning, a method that achieves unlearning by modifying the output logits. It leverages a pair of small, white-box models—identically initialized—to compute a logit offset, defined as the difference in their output logits for the same input. One model is fine-tuned with an unlearning objective (e.g., GA), while the other remains frozen. At inference time, the learned offset is added to the logits of the gray-box LLM, effectively steering its output away from sensitive content. This approach enables plug-and-play unlearning, but it requires the white-box models to share the same tokenizer as the target model and to have access to its output logits for modification.

Ji et al.~\cite{ji2024reversing} propose an unlearning framework that inverts the usual forget-retain objective. Instead of modifying the target LLM directly, they train an assistant model to memorize forget data and forget retain data. Unlearning is achieved by subtracting the assistant's logits from the target's:
\begin{align*}
    l_f(y \mid x)=l_{\text {target }}(y \mid x)-\alpha \cdot l_{\text {assistant }}(y \mid x)
\end{align*}
When the target model and assistant model both assign high probability to a token (e.g., a correct answer from the forgetting data), the subtraction significantly reduces its score—effectively suppressing it.
When the assistant model assigns low or uniform scores (as trained) on retain data, the subtraction has minimal effect—preserving the target model’s original predictions.
This contrastive behavior cancels the knowledge learned from specific forgetting data without changing the original LLM’s parameters. 

\subsubsection{Retrieval-based} \label{sec:retrieval_based} This method is similar to RAG-based unlearning, but with a key difference: the retrieved content is used to block the generation of specific answer tokens~\cite{deng2025guard}. The method first employs an MLP classifier to determine whether an input prompt targets forgertting data. If so, it retrieves the most semantically similar answer from the forgetting set and extracts key phrases as forbidden tokens. During decoding, it dynamically penalizes or blocks candidate tokens using both token-level hard matching and soft semantic matching, effectively suppressing forgotten content while preserving model fluency and utility.




\section{Do the removal-intended methods truly remove forgetting data?}
\label{sec:truly}

In the proposed taxonomy, removal-intended methods—primarily GA-based approaches—directly modify the model parameters $\theta$ to erase the impact of the forgetting data. Their unlearning effectiveness has been shown in various studies~\cite{zhang2024negative, yao2023large} and benchmarks~\cite{maini2024tofu, shi2024muse}, where unlearned models successfully avoid reproducing the forgetting content in generated responses.

However, no theoretical framework currently guarantees that GA-based methods can truly remove the influence of data from LLMs, nor is there direct empirical evidence supporting this claim. The only guidance is the initial intuition of GA: \textit{by fine-tuning with a reversed training loss of LLM, GA can negate the training influence of }$\mathcal{D}_f$. 
Recently, this assumption has been increasingly questioned. 
Several studies suggest that GA-based methods may only reflect superficial changes in model behaviors instead of truly removal~\cite{thaker2024position, lynch2024eight}, raising an important question: \textit{do these methods truly remove forgetting data?} Although GA-based methods are designed to achieve the true removal, emerging evidence indicates that they may, in practice, only perform suppression~\cite{thaker2024position, ren2025general}.

In this section, we first discuss the evidences verifying the existence of this problem (Section~\ref{sec:ga_fail}). Then we show a new understanding perspective to explain the mechanism of GA-based methods and why they do not truly remove (Section~\ref{sec:ga_understand}). Finally, we discuss the open problems: Can we truly remove? What is the challenge? Do we need true removal or is behavioral
suppression sufficient? (Section~\ref{sec:open_problem})

\subsection{Failures of removal-intended unlearning}
\label{sec:ga_fail}



Recent research examining the capabilities of GA-based unlearning reveals two key insights. First, unlearned models can still produce outputs related to the forgetting data under adversarial attacks~\cite{schwarzschild2024rethinking, yuan2025towards, to2025harry}, suggesting that the removal may be superficial and fails to eliminate deeper representations of the forgotten knowledge. Second, unlearned models remain highly vulnerable to subsequent fine-tuning: even without access to the forgetting content, these models can re-memorize it~\cite{hu2024jogging, deeb2024unlearning}. This indicates that hidden patterns associated with the forgetting data may still persist—temporarily suppressed by unlearning but easily reactivated through additional training. In this subsection, we expand on these two observations.

\subsubsection{Adversarial attacks} 
GA performs well on the widely used QA benchmark TOFU~\cite{maini2024tofu}. When prompted with a question $x$ from the forgetting set $\mathcal{D}_f$, the unlearned LLM typically hallucinates the answer rather than recalling the correct response. However, Schwarzschild et al.\cite{schwarzschild2024rethinking} observe that if the model is instead prompted with a subsequence of $x$, the GA-unlearned model can still generate the correct answer $y$. This suggests that the model may memorize and suppress the exact full sequence of $x$, while partial cues still activate the forgotten knowledge. Similarly, Yuan et al.\cite{yuan2025towards} apply the GCG attack~\cite{zou2023universal} to optimize adversarial suffixes and find that 54\% of the forgetting knowledge can be recovered—despite having no access to model parameters. To et al.~\cite{to2025harry} also use the GCG attack to assess whether the model has truly forgotten the data. They find that larger LLMs are more vulnerable to knowledge extraction attacks. The authors further evaluate Task Arithmetic, another removal-intended method, and find it poses even higher risk of forgetting data recovery than GA-based methods. This may be because model editing via task arithmetic is more aggressive, and the need to constrain the editing magnitude to preserve utility reduces its unlearning effectiveness.

To provide more evidence, Hong et al.~\cite{hong2024dissecting} design a series of activation patching and parameter restoration experiments. They find that fine-tuning-based unlearning mainly manipulates behavior through output-layer coefficients, rather than actually removing stored knowledge.

In addition, Soft token attacks are proposed to bypass safety alignment and unlearning in open-source LLMs by directly perturbing the continuous input embeddings rather than discrete tokens~\cite{schwinn2024soft, du2024textual, zou2024improving}. For example, Schwinn et al.~\cite{schwinn2024soft} add a adversarial embedding $e_i^{a d v}$ after a sequence $x$ as its embeddings. It is optimized to minimize the cross-entropy loss between the model’s output and $x$'s groundtruth response:
\begin{align}
    \mathcal{L}_{\text{CE}}\left(f\left(x \| e_i^{a d v}\right), y_i\right)
    \label{eq:sta}
\end{align}
Experiments show this approach outperforms prior discrete attacks in success rate, efficiency, and its ability to recover supposedly deleted or pretraining data.

Interestingly, Chen et al.~\cite{chen2025soft} find that soft token attacks cannot reliably audit unlearning. This attack is too powerful and can generate the knowledge even the model has never seen. Soft token attacks function more like an extreme case of prompt tuning. It optimizes the embedding space to enable the model to output nearly arbitrary sequences. This means the forgetting knowledge is injected by the adversarial embeddings and lead to false positive.

\subsubsection{Fine-tuning after unlearning} It is found that the unlearned model might recover the forgetting knowledge after fine-tuning with other data~\cite{hu2024jogging, deeb2024unlearning}. \cite{hu2024jogging} reveals that models can easily recover supposedly unlearned information through small-scale finetuning on related but benign data—highlighting that unlearning is often superficial. \cite{deeb2024unlearning} introduces a more rigorous evaluation framework, showing that state-of-the-art unlearning methods often fail to remove information from model weights, as the forgotten knowledge can be recovered through retraining on independent facts. These works demonstrate that existing unlearning approaches tend to hide knowledge rather than truly erase.

\begin{tcolorbox}[colback=gray!20, 
                  colframe=black,
                  coltitle=black,
                  colbacktitle=gray!20,
                  boxrule=0.4pt,
                  arc=4pt,
                  fonttitle=\itshape,
                  left=6pt, right=6pt, top=6pt, bottom=6pt,
                  enhanced]
\textbf{Takeaway.} GA-based unlearning is often superficial and often fails to truly remove: forgetting knowledge can resurface under adversarial prompts or reappear after benign fine-tuning, revealing that current methods suppress rather than erase internal representations.
\end{tcolorbox}

\subsection{A new understanding: removal-intended methods are actually doing suppression}
\label{sec:ga_understand}

Based on the above observations, researchers have begun to realize that simply applying a reversed loss may not truly eliminate the training traces of the forgetting data. This has prompted a search for a deeper understanding of how GA-based methods actually operate. In this subsection, we present two key phenomena—\textit{over-generalization} and \textit{syntactic similarity}—that inspire a new understanding of GA-based unlearning. We analyze these phenomena and discuss how they contribute to a revised understanding of the underlying mechanisms of GA-based methods.

\textbf{From over-generalization to unlearning signals.}
People find that the unlearning will \textbf{over-generalize} to other data that is related to forgetting knowledge~\cite{zhang2024safe}. When models are trained to forget certain knowledge (e.g., theft), they also implicitly forget similar contents (e.g., bomb-making). 
In the beginning, people connect this overgeneralization with the explanation of the reduction of utility (collateral damage)~\cite{patil2025upcore, maini2024tofu, jeung2025dusk}. If GA-based methods truly negate the training influence, over-generalization would suggest they also suppress related knowledge. However, since GA often only hides the forget data, over-generalization instead implies that these methods may unintentionally suppress any data similar to the forgetting data.

While the above seems just a problem of unclear decision boundary, Thaker et al. put forward a novel perspective in their position paper~\cite{thaker2024position}. They find that when adding a question of forgetting data before/after a question of retaining data, the unlearned would be unable to answer the retaining data, too. This is a novel evaluation method because previous works only test the forgetting and retaining data separately, but they do it simultaneously. 


Based on this observation, the unlearning mechanism of GA is further explained as the unlearning signal by~\cite{ren2025general}. In \cite{ren2025general}, the authors observe some properties in the representation space $h(x)$. 
They provide three sub-sets: ``forgetting'' data, ``retaining'' data and ``never-seen'' data. They have similar representations because they are all synthetic data sampled from the same distribution of TOFU benchmark (details of TOFU will be discussed in Section~\ref{sec:benchmarks}). However, after unlearning, in the representation space, the clusters of ``retaining'' data and ``never-seen'' data are still close, while the cluster of ``forgetting'' data is far from them. This means unlearned LLM still recognize the forgetting data. The unlearned model does not forget it, but distinguish it from other data. 
The authors further find that better unlearning effectiveness is likely to be associated with better distinction. 
Lastly, they also add forgetting data questions into normal (retaining/never-seen) data questions and find that once mixed with target data, the representations of normal data is dominated by forgetting data (which is pulled toward the distinct cluster of forgetting data). Consequently, the model’s ability to answer normal questions deteriorates. As the authors conclude, this implies that, instead of removing the forgetting data, GA-unlearned models treat it as a unlearning signal to suppress the generation. They behave like unlearning once there is a unlearning signal in the prompt. Worth to noted that this is actually the same as what suppression-intend unlearning is doing.

The previous discussion provides a refined understanding of the underlying behavior of GA-based unlearning methods. However, a critical question remains: why are these methods inherently limited in their effectiveness? The following paragraphs offer preliminary explanations based on syntactic similarity and related studies.

\textbf{From syntactic similarity to true intention.}
The second phenomenon is the role of syntactic similarity, which is highly related to the selective forgetting in GA-based methods in Section~\ref{sec:selective_forgetting}. This phenomenon reflects the limited intention behind GA loss. 
Specifically, Chang et al~\cite{chang2025retain} find that syntactic similarity plays a critical role in maintaining model utility. Syntactically similar means having the same sentence structure or grammatical pattern. For example, ``\textit{Who is the author of Watermelon on the Moon?}'' and ``\textit{Who is the author of Attention is all you need?}'' are syntactically similar (the same sentence structure and different literature titles). They find that more similar syntactic between forgetting data and retaining data would have a better model utility. When GA optimizes forgetting loss on the benign tokens (the tokens that are not directly related to the forgetting data such as ``\textit{Who is the author}''), these benign tokens are also calculated in retaining loss, which relieves the destroy on the model utility for benign tokens.

This indicates that GA does not really negate the knowledge.
As shown by \cite{ren2025keepingeyellmunlearning} if we look at the GA loss (Eq.~\ref{eq:ga_init}), what the objective actually does is: \textbf{if $x$ is in the prompt, increase its output error}. GA loss would not distinguish benign tokens or forgetting tokens (tokens that are directly related to forgetting data). The Gradient Ascent on benign tokens are meaningless and even harmful for the model. This explains the phenomenon of syntactic similarity.
If there are corresponding syntactically similar tokens in retaining data, the destruction on benign tokens can be make up to some extent. This also explains the mechanist of unlearning signal: if forgetting data is in the prompt, increase its output error (i.e., behave like know nothing about it).

Although GA intends to remove the knowledge from LLMs, it actually does similar things as suppression-intended unlearning. However, we should not ignore the significant contribution of GA at this stage of LLM unlearning. Although people haven't fine a better way to truly remove the knowledge, we believe GA will be the basic for the future work. For example, as we mentioned in Section~\ref{sec:selective_forgetting}, selective forgetting has been proposed to explore a correct way to using GA.

\begin{tcolorbox}[colback=gray!20, 
                  colframe=black,
                  coltitle=black,
                  colbacktitle=gray!20,
                  boxrule=0.4pt,
                  arc=4pt,
                  fonttitle=\itshape,
                  left=6pt, right=6pt, top=6pt, bottom=6pt,
                  enhanced]
\textbf{Takeaway.} GA-based unlearning does not truly remove knowledge; instead, it treats the forgetting data as a signal to trigger unlearning behavior. When this signal is detected, the model simulates ignorance. This aligns with the design of GA loss, which increases the model's output error when presented with the forgetting data.
\end{tcolorbox}

\subsection{Open problems}
\label{sec:open_problem}
After reviewing and discussing the existing research in removal-intended unlearning, several important questions remain unresolved. In this subsection, we identify some key open problems.

\textbf{Can we truly remove the influence of forgetting data?}
The fundamental goal of removal-intended unlearning is to erase the influence of specific data $\mathcal{D}_f$ from model parameters $\theta$. However, no theoretical framework currently guarantees such removal in LLMs, nor are there conclusive empirical methods to verify it. The key challenge is the entangled, non-linear way in which knowledge is stored across millions or billions of parameters in transformer-based models. Simply reversing the loss signal, as done in GA, may not undo the distributed effects of training.

To achieve true removal, we may need new theoretical tools to (1) trace the specific parameter regions influenced by $\mathcal{D}_f$ and (2) isolate and invert those changes without degrading overall utility. This will likely require advances in interpretability, causal attribution, or model editing beyond current fine-tuning paradigms. Moreover, new benchmarks and evaluation protocols are needed to distinguish true removal from behavioral suppression, especially under adversarial or compositional prompts.

\textbf{Do we need true removal, or is behavioral suppression sufficient?}
From a practical standpoint—especially in commercial or production-level applications—strictly enforcing true removal may not always be necessary. In many deployment scenarios, behavioral suppression may be sufficient as long as the model no longer reproduces the target knowledge under expected usage conditions. Instead of pursuing theoretical perfection, practitioners should focus on whether the model meets practical requirements. If suppression-intended unlearning aligns with practical requirements, it can even be prioritized over true removal.

However, this should not discourage ongoing research into true removal. The pursuit of actual knowledge removal remains scientifically valuable, as it can inspire deeper understanding of fine-tuning, alignment, interpretability, and model editing. Developing mechanisms to trace, attribute, and eliminate specific knowledge in LLMs can open up broader research directions, including modular training, causal reasoning, and long-term memory control. Therefore, while suppression may be acceptable in practice, removal should remain a central research goal.

\section{Evaluation metrics and benchmarks}
\label{sec:eval}

Evaluation plays a critical role in the development of LLM unlearning. A reasonable and accurate evaluation can help researchers design more effective methods. At a high level, unlearning evaluation encompasses two key aspects: \textit{unlearning effectiveness} and \textit{model utility}. Numerous metrics have been proposed, and several benchmarks have been established to assess these dimensions. In this section, we review existing metrics and benchmarks, and discuss their limitations along with potential future directions.

\subsection{A review of existing metrics}

Unlearning effectiveness measures whether a model continue to output or leak information related to the forgetting data and knowledge, while model utility evaluates the extent to which the model retains its normal functionality on non-forgotten tasks. This subsection first introduces three general metrics applicable to both aspects, followed by metrics specifically tailored to unlearning effectiveness.

\subsubsection{General metrics} Traditional metrics for generation quality and accuracy can be used to measure both unlearning and utility, we first present this kind of metrics below:

\begin{itemize}[leftmargin=1.5em]
    \item \textbf{ROUGE Recall.}~\cite{maini2024tofu, shi2024muse, yuan2024closer} measures the overlap between the model's generated answer and the ground truth at the word level. Specifically, it evaluates how much of the ground truth answer is recalled in the output. The precision of ROUGE is usually not used because the generation usually contains additional or auxiliary content beyond the ground truth answer, which should not influence the measurement.
\item \textbf{Probability.} This metric~\cite{maini2024tofu, ren2025general, yuan2024closer} measures the likelihood that the model reproduces the ground truth answer, typically evaluated token by token.
    \begin{align*}
        Prob(y \mid x)=\frac{1}{T} \sum_{t=1}^T \mathrm{p}\left(y_t \mid x, y_{<t}\right)
    \end{align*}
    A high value indicates that the model still assigns high confidence to the answer.
    \item \textbf{Multi-choice accuracy.}~\cite{liwmdp, huu2025improving, ren2025general} evaluates whether the model can correctly select the ground truth answer from a set of candidate options. It reflects the model’s ability to distinguish the correct answer from distractors in a multiple-choice setting.
\end{itemize}

\subsubsection{Unlearning effectiveness metrics} Some metrics have been proposed or adapted to evaluate unlearning effectiveness. We begin by introducing two categories of methods that uses existing tools:

\begin{itemize}[leftmargin=1.5em]
    \item \textbf{Membership Inference Attacks (MIA)} are used by some benchmarks to evaluate the privacy leakage~\cite{shi2024muse, jin2024rwku, ramakrishna2025lume}. If the training data can be successfully inferred, it means the unlearning effectiveness may be not resistant. The common MIA includes LOSS~\cite{yeom2018privacy}, Zlib Entropy~\cite{carlini2021extracting}, Min-K\% Prob~\cite{shi2023detecting} and so on.
    \item \textbf{Robustness} is an important evaluation aspect. As discussed in Section~\ref{sec:ga_fail}, it serves as a key tool for exposing superficial unlearning behaviors. A common approach involves using adversarial prompts to test the robustness of the model~\cite{jin2024rwku, yuan2025towards, to2025harry}.
    
\end{itemize}

In addition to these two methods that leverage existing tools, several new tools have also been developed to measure unlearning effectiveness:

\begin{itemize}[leftmargin=1.5em]
    \item \textbf{Watermark.} Lu et al.~\cite{lu2025waterdrum} propose to embed imperceptible, owner-specific watermarks into training data and verifies their presence in model outputs after unlearning.
    \item \textbf{Gradient Effect.} Want et al.~\cite{wang2025rethinking} propose Gradient Effect that quantifies how an unlearning objective impacts model performance via gradient alignment. Specifically, it computes the dot product between the gradients of the unlearning objective $\mathcal{L}_{\text{u}}$ and the risk function $\mathcal{R}$.
    \item \textbf{Unlearning Shapley}. Ma et al.~\cite{ma2025losing} propose a novel data valuation framework that combines machine unlearning with Shapley value theory~\cite{shapley1953value}. It measures the performance drop after unlearning target data from a pretrained model to estimate its value.
    \item \textbf{Representation-level.} Xu et al.\cite{xu2025unlearning} introduce a set of representation-level evaluation metrics for unlearning. These metrics go beyond token-level accuracy or perplexity to assess whether the model's internal feature representations are genuinely altered or merely superficially perturbed.
\end{itemize}

\subsection{A review of existing benchmarks} \label{sec:benchmarks}

Benchmarks are essential for evaluating unlearning methods, offering controlled and synthetic settings for fair comparison.A critical aspect is the construction of forgetting data, which directly affects evaluation validity. As unlearning methods continue to evolve, several benchmarks have been proposed to match this progress. In this section, we introduce a few representative and widely used benchmarks.


\textbf{TOFU}~\cite{maini2024tofu}. TOFU uses a synthetic dataset of fake books and authors. The data corpus is a set of QA pairs. Since no LLM has trained on the synthetic dataset, the tested model has to be fine-tuned to learn from the dataset first. Then the synthetic dataset is separated into forgetting data and retaining data. The utility is tested by three QA sets in TOFU: retaining data, and two real QA sets (real authors and world knowledge). The advantage of fine-tuning is easy to control, and the new knowledge has not much entanglement with real data. A few new benchmarks are also proposed based on it like PerMU focusing on robustness~\cite{wang2025erasing} and R-TOFU focusing reasoning~\cite{yoon2025r}.

\textbf{MUSE}~\cite{shi2024muse}. MUSE has a similar pipeline as TOFU. It first fine-tunes LLMs to learn forgetting data. But it does not use synthetic datasets. It uses a dataset that is not in the training data of the tested LLMs. The advantage of this benchmark is that it provides a more comprehensive metrics, especially the test on scalability and sequential unlearning.

\textbf{WMDP}~\cite{liwmdp}. WMDP does not fine-tune to learn new knowledge. It tests the unlearning of harmful bio and chemical knowledge. It uses LLMU~\cite{hendrycks2020measuring} and MT-Bench~\cite{zheng2023judging} to test the model utility across a broad spectrum of subjects.
A multi-choice format is used to test the unlearning effectiveness and utility.

\textbf{RWKU}~\cite{jin2024rwku}. RWKU targets 200 real-world famous entities and rigorously evaluates unlearning effectiveness using fill-in-the-blank, question-answer, and nine types of adversarial probes. It also assesses side effects on neighboring knowledge and general utility.

In summary, the choice of benchmark significantly influences how unlearning methods are evaluated, but no single benchmark is inherently superior to others. Each focuses on different aspects—such as control, realism, scalability, or robustness—and over-reliance on a specific benchmark may lead to models being optimized for test performance rather than real-world forgetting objectives. Therefore, it is important to encourage the development of diverse and comprehensive benchmarks to ensure a more robust and generalizable assessment of unlearning methods.

\subsection{Toward better evaluations: challenges and considerations}
\label{sec:better_eval}

To foster more robust and meaningful evaluations, we outline several key considerations that should guide future efforts. These include both methodological principles and practical constraints that are often overlooked in current benchmarking practices.


\textbf{Fixed and greedy sampling is too narrow.} 
Scholten et al.~\cite{scholten2024probabilistic} arguing that standard deterministic (greedy) evaluations fail to capture real-world risks such as information leakage in unlearning and alignment tasks. Although sometimes a model fails to generate a forgotten answer, the model still assigns a high probability to the correct answer, indicating that the knowledge is internally retained~\cite{krishnan2025not}. Thus, relying solely on fixed or greedy decoding may underestimate the model's actual knowledge retention; probabilistic evaluation across a distribution of outputs provides a more realistic and robust picture.



\textbf{Unclear settings: what should the user submit?}
Current unlearning scenarios are all based on benchmark assumptions, where the data to be unlearned is fixed and clearly defined. However, in real-world applications, what exactly should users provide? Should they specify abstract concepts or concrete corpora? Abstract concepts are often too vague and underspecified, while user-provided corpora lack standardization and may vary in quality. We attempt to answer this question through real-world case studies, but unfortunately, although various unlearning methods have been proposed, there is no commercial product that explicitly offers unlearning capabilities or claims to have integrated unlearning techniques. As a result, there is still no practical and reliable answer to this question.

\textbf{Fine-tuning knowledge vs. pre-training knowledge.}
For TOFU and MUSE, the model first fine-tunes to learn new knowledge. The new knowledge is separate and easiser to control than pre-training knowledge. However, fine-tuning would also bring two disadvantages. The first is that fine-tuning suffers from catastrophic forgetting~\cite{luo2023empirical}. When unlearning on the fine-tuned models, the reduce of the ability in answering forgetting data might be somehow attributed to catastrophic forgetting, making it hard to isolate the true effect of unlearning. The second is that fine-tuning on a small synthetic dataset would lead to overfitting and will cause the reduction of utility on two real QA sets~\cite{maini2024tofu}. Thus, the reduction of utility bring by unlearning is hard to quantify.

\textbf{Core-set vs. complete set.}
Patil et al.~\cite{patil2025upcore} propose that unlearning the whole forgetting set might be one of reason of the reduction in model utility. Thus, they identify and prune 10\% to 30\% outliers from the forget set. Pal at al.~\cite{pal2025llm} further study this core-set effect. The authors find that even a randomly selected 5\% subset can result in comparable unlearning performance to using the full dataset. These findings indicate that existing evaluations may overestimate the necessity of forgetting the entire dataset. Thus, future unlearning benchmarks should consider core-set-aware evaluation protocols to more realistically assess unlearning effectiveness and efficiency.

\textbf{What is a reasonable retaining set?}
Syntactic similarity has already been shown to affect utility degradation—when the retaining set is more similar to the forgetting set, the utility drop tends to be smaller~\cite{chang2025retain}. Some benchmarks use idealized retaining sets; for example, TOFU selects retaining data that is highly similar to the forgetting data. However, such assumptions are unrealistic in real-world scenarios, where we often do not know exactly what data the user wants to unlearn in advance, making it difficult to provide such well-matched retaining sets. Therefore, benchmark design should take this limitation into account when selecting the retaining set.



\textbf{Overlooked utility evaluation.} Utility evaluations based solely on unlearning benchmarks may fail to reflect the actual performance of the LLM. In practice, every LLM undergoes internal performance testing before being released or deployed, and utility should be assessed using these evaluation suites. For example, ChatGPT has its own benchmark suite~\cite{achiam2023gpt}, and unlearning should be considered acceptable only if the model’s performance on these internal tasks remains stable. However, current unlearning benchmarks often use their own utility criteria, which may not align with real-world expectations.

\section{Other gaps in existing unlearning}
\label{sec:gaps}

To enable unlearning techniques to be truly adopted in real-world applications, this section discusses the key gaps that still hinder practical deployment. Some of these challenges remain open problems and have yet to receive sufficient attention from the community. We highlight these issues in the hope that future research will address them, ultimately bridging the gap between unlearning research and its real-world implementation.

\textbf{Model utility.} The harm to model utility is always the most critical obstacle to deployment. Let's think from the perspective of a model builder: if unlearning did not bring any negative impact on model performance, who would mind adding such a beneficial feature to their LLM? Therefore, regardless of the unlearning effectiveness, model utility must be preserved to give model builders confidence in using it. Thus, both maintaining and verifying utility are of paramount importance.

\textbf{Sequential unlearning.} While preserving utility is a well-recognized goal in unlearning, one particularly challenging setting is sequential unlearning, where utility degradation becomes more severe and difficult to mitigate. In real-world applications, users are likely to submit new unlearning requests continuously, and most existing methods suffer from accumulated reductions in model utility as a result.

There are currently two main ideas for addressing this challenge. One approach is to avoid sequential unlearning altogether and instead perform unlearning on the entire set of forgetting data each time. The coreset effect suggests that scalability is promising—meaning we may not need to process the full dataset to achieve good forgetting performance. However, if we persist with sequential unlearning and unlearn one subset at a time, the utility degradation quickly becomes significant. For example, even if each step results in just a 3\% drop in utility, after 20 iterations, the model could lose up to 45\% of its utility. In practice, users are likely to submit far more than 20 unlearning requests, and in fact, the utility loss per step is often much greater than 3\%~\cite{maini2024tofu, shi2024muse}, especially for methods that directly modify model parameters or internal representations.

An alternative and promising direction is to base unlearning on RAG as we discussed in Section~\ref{sec:rag_based}. New requests only need to update the retrieval system.

\textbf{Scalability of model size.}
Speaking of scalability, another important aspect to consider is the scalability with respect to model size. Most current studies conduct experiments on relatively small models, such as 7B or 8B parameters~\cite{maini2024tofu, liwmdp, jin2024rwku}. However, for real-world deployment, it is essential to evaluate unlearning methods on much larger models. This shift introduces entirely new challenges, both in terms of computational resources and in the design of scalable and efficient unlearning methods that can handle the complexity of large-scale LLMs.

\textbf{Poisoning risks in unlearning submissions.}
If the unlearning service accepts user-provided corpora, as discussed in Section~\ref{sec:better_eval}, the quality of such data is often difficult to guarantee. Taking this a step further, what if someone intentionally injects poisoned data into the unlearning request? How can we ensure that the data is clean and trustworthy? In fact, recent work has demonstrated that it is possible to poison the unlearning data to deliberately degrade the model's utility~\cite{ren2025keepingeyellmunlearning}. Therefore, ensuring that user-provided corpora are both safe and standards-compliant is of critical importance.

\textbf{Interference Between Different Training and Inference Stages.}
Can unlearning interfere with other training processes such as RLHF? It is currently assumed that unlearning is performed only after the model has been fully trained, but does this risk undermining the alignment achieved through RLHF? Many studies suggest that alignment is rather superficial~\cite{wei2024assessing, lin2023unlocking}. If further RLHF tuning is required after unlearning, it may compromise the intended purpose of unlearning. Similarly, if downstream tasks are introduced after unlearning, would that diminish the effectiveness of unlearning? Moreover, if quantization is applied during deployment, the effectiveness may also be reduced~\cite{zhang2024catastrophic}—this is another factor that needs to be considered.

\section{Conclusion}

Machine unlearning has emerged as a critical technique to address growing concerns around privacy, copyright, and regulatory compliance in LLMs. This paper presents a structured overview of unlearning in LLMs. While prior work focuses on technical methods, we propose a new taxonomy based on intentions: removal vs. suppression.
Beyond taxonomy, we make three core contributions: (1) we examine whether popular removal-intended methods like GA can truly erase knowledge, revealing both theoretical and empirical limitations; (2) we survey evaluation strategies, identifying key shortcomings in current metrics and benchmarks; and (3) we highlight open challenges for real-world deployment, including usability, side-effect mitigation, and verification. This study aims to provide a principled foundation for future research and practical deployment of unlearning in LLMs.

\newpage

\bibliographystyle{IEEEtran}
\bibliography{reference}
%



\end{document}